\ifdefined\pdfminorversion
\fi
\documentclass[10pt,twocolumn,letterpaper]{article}
\usepackage[letterpaper,margin=1in,columnsep=0.33in]{geometry}
\usepackage{times}
\usepackage{graphicx}
\usepackage{amsmath}
\usepackage{amssymb}
\usepackage{booktabs}
\usepackage{multirow}
\usepackage{bm}
\usepackage{mathtools}
\usepackage{enumitem}
\usepackage{caption}
\usepackage{xcolor}
\makeatletter
\@ifundefined{c@algorithm}{\newcounter{algorithm}}{}
\makeatother

\definecolor{linkblue}{rgb}{0.21,0.49,0.74}
\usepackage[breaklinks,colorlinks,allcolors=linkblue]{hyperref}

\title{QueenVIS: Rethinking Image-Only Training for Video Instance Segmentation via Query Enrichment}

\author{
	Arian Kheirandish$^1$ \and
	Fardin Ayar$^1$ \and
	Ehsan Javanmardi$^2$ \and
	Manabu Tsukada$^2$ \and
	Mahdi Javanmardi$^{1*}$ \\
	$^1$Amirkabir University of Technology (AUT), Tehran, Iran \\
	$^2$The University of Tokyo, Tokyo, Japan \\
	{\tt\small $^*$mjavan@aut.ac.ir}
}
\date{}

\begin{document}
	
	\maketitle
	
\begin{abstract}
Video instance segmentation (VIS) requires models to detect, segment, and track object identities across frames, and most methods enforce temporal consistency through video-level supervision. Image-only training approaches, with MinVIS as one prominent example, have challenged this assumption, reaching competitive VIS without video training by treating frames as independent images and associating instances only at inference. The field has nonetheless moved toward ever more elaborate video-trained trackers, which depend on costly identity-consistent annotations, leaving the image-only direction under-explored. A diagnostic analysis identifies object query quality as the bottleneck: queries trained only to localize objects within a frame drift apart across frames and destabilize association. QueenVIS introduces a query-centric framework for strengthening image-trained VIS. During single-frame training, we enrich Mask2Former queries with two auxiliary heads: a feature-prediction loss that aligns each query with the pooled backbone descriptor of its instance, and a center-prediction loss that injects spatial structure. Both heads are discarded at inference, adding zero parameters, and temporal identity is maintained by a training-free query-propagation and memory-bank scheme. On YouTube-VIS and OVIS with a ResNet-50 backbone, QueenVIS improves over MinVIS, up to +6.7 AP on YouTube-VIS, +4.8 AP on OVIS, and +10.3 AP on the long-sequence YouTube-VIS split. QueenVIS achieves 50.9 AP on YouTube-VIS and remains competitive with recent video-supervised state-of-the-art, without processing a single video clip during training. Our findings suggest that strengthening the discriminative power and temporal stability of object queries is an important, underexplored axis for VIS.

\noindent Code and models: \url{https://github.com/ArianKheir/QueenVIS}
\end{abstract}
	
	\begin{figure}[!tb]
		\centering
		\includegraphics[width=0.85\linewidth]{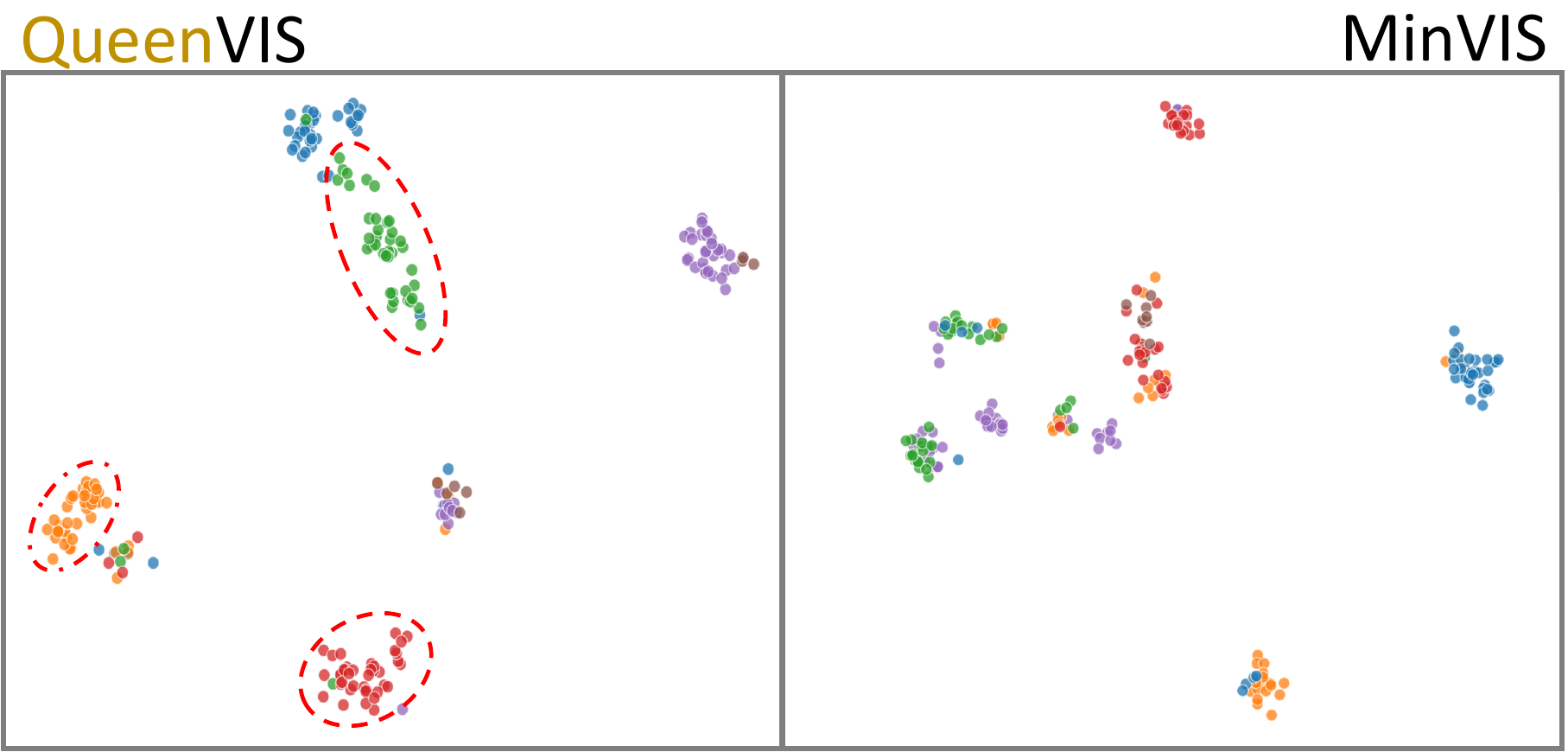}
		\caption{t-SNE visualization of object queries on the OVIS dataset. Each colored point represents a query for a ground-truth identity, matched across QueenVIS (left) and MinVIS (right). Red dashed ovals explicitly denote instances where QueenVIS successfully maintains tight clusters compared to the widely scattered baseline. While MinVIS suffers from severe feature drift that destabilizes association, QueenVIS embeds both spatial and appearance priors to keep instance queries highly cohesive across frames.}
		\label{fig:tsne}
	\end{figure}	
	
	\section{Introduction}
	\label{sec:intro}
	Video instance segmentation (VIS) extends image instance segmentation to the temporal domain, where the per-frame masks of each object must additionally be linked into a single identity that persists throughout a video. Its main difficulty is keeping these identities consistent despite large appearance changes, heavy occlusion, camera motion, and object deformation. VIS is also a building block for many applications, such as autonomous driving, robotic manipulation, and long-term video understanding.

	The common approach to these challenges is explicit temporal modeling. Building on the transformer query formulation of DETR~\cite{carion2020detr} and Mask2Former~\cite{cheng2021mask2former}, methods such as SeqFormer~\cite{wu2022seqformer}, IDOL~\cite{wu2022defense}, and VISAGE~\cite{kim2024visage} train on multi-frame clips with identity losses, temporal attention, and dedicated tracking modules. These methods reach strong temporal accuracy, but at a cost. Video-level training needs temporally linked, identity-consistent annotations that are far costlier to collect than independent image masks, and it raises training cost because the model must process several frames at once rather than one. This burden is heaviest for offline methods, which run on long clips or whole sequences and therefore demand large memory and compute. Together, these requirements limit scalability.

	MinVIS~\cite{huang2022minvis} observed that queries trained only to discriminate instances \emph{within} a frame are already temporally consistent enough to be tracked by bipartite matching alone, with no video-level supervision. This consistency, however, emerges as a byproduct rather than an objective: queries encode only what intra-frame separation demands, which proves brittle under the appearance change and occlusion that characterize real video. Perhaps for this reason, most subsequent work has retreated to the video-supervised regime: GenVIS~\cite{heo2023generalized}, TCOVIS~\cite{li2023tcovis}, and DVIS~\cite{zhang2023dvis} layer increasingly elaborate temporal modeling on multi-frame training, while VISAGE~\cite{kim2024visage} and CAVIS~\cite{lee2025cavis} enrich instance representations with appearance and context under the same supervision. Both treat video as the source of the missing signal. We argue instead that much of this signal is already latent in per-frame features and can be injected into the object query at training time, with no video clips required. With only per-frame classification and mask losses, queries learn to localize but not to re-identify, so the embeddings of a given instance diverge across frames, as Figure~\ref{fig:tsne} illustrates for MinVIS.

	Building on this observation, we propose QueenVIS (QUEry ENrichment), a framework that enriches object queries with appearance and spatial cues under image-only supervision. Much of the appearance cue that video-supervised methods extract through temporal modules is already present in the per-frame backbone features, and need not be learned from video. We recover it with a lightweight MLP head that, during training only, regresses each query toward the pooled backbone descriptor of its matched instance, transferring this appearance signal into the query. Appearance alone, however, cannot disambiguate visually similar instances that lie in different parts of the frame, a common source of long-range identity switches. We therefore also predict each instance's center with a single linear layer, injecting a spatial cue that keeps such instances separable, an idea already validated for video panoptic segmentation~\cite{ayar2024lidar}. Both heads are discarded at inference and add no parameters. To associate the resulting queries over time without a learned tracker, we adopt a fully training-free scheme: high-confidence content queries from preceding frames are blended into the current frame as a warm start, and an off-the-shelf memory bank~\cite{kim2024visage} supports bipartite matching over recent frames.

	Our core contributions are summarized as follows:
	\begin{itemize}[leftmargin=1.5em]
		\item We show that image-only training can be far more competitive than commonly assumed: enriching object queries with dense appearance and spatial priors during image-only training closes much of the gap to video-supervised models.
		\item We introduce two complementary training-only objectives, feature prediction and center prediction, that adapt Mask2Former queries for cross-frame association without any video-clip supervision. Both heads are discarded after training, so QueenVIS adds zero parameters and matches the MinVIS inference cost (identical GFLOPs), sidestepping the heavy compute and identity-consistent annotation pipelines that video-supervised trackers require.
		\item We propose a fully training-free, confidence-guided query-propagation mechanism for cross-frame association, and combine it with a memory bank~\cite{kim2024visage} to improve identity consistency under severe occlusion, all without any tracking parameters or video supervision.
		\item With a ResNet-50 backbone, QueenVIS improves over MinVIS across YouTube-VIS 2019/2021 and OVIS without any video clips, matching or surpassing several video-supervised methods while narrowing the gap on the heavily occluded OVIS, and the margin grows largest on the long-sequence YouTube-VIS 2022 split.
		\item Using the model-agnostic oracle framework of Hamdi et al.~\cite{hamdi2026mind}, we attribute this gain to association, shrinking the tracking-recall gap to essentially the oracle ceiling, tighter than even video-supervised trackers achieve.
	\end{itemize}

	These results suggest the image-only regime is far from exhausted: stronger query optimization alone closes much of the gap to video-supervised methods while avoiding their costly multi-frame annotation and temporal training, offering a practical path to competitive VIS when labeled video is scarce or compute is limited.

	\section{Related Work}
	
	\subsection{Query-Based Object Detection and Segmentation}
	Our method builds on the query-based segmentation paradigm. DETR~\cite{carion2020detr} reformulated object detection as direct set prediction, using bipartite matching and a fixed set of transformer object queries to remove non-maximum suppression and anchor generation. MaskFormer~\cite{cheng2021maskformer} extended this formulation to segmentation by having queries jointly predict class labels and mask embeddings, and Mask2Former~\cite{cheng2021mask2former} added masked attention, which confines cross-attention to predicted mask regions and set the standard for image-level panoptic and instance segmentation. In all of these methods, each query is optimized only to separate instances within a single image. QueenVIS keeps the Mask2Former architecture unchanged but reshapes how its queries are trained, so that they encode each object's appearance and spatial priors rather than representations optimized only for within-frame instance separation.
	
	\subsection{Video Instance Segmentation}
	Most VIS methods reach temporal consistency through video-level supervision. Early tracking-by-detection pipelines~\cite{yang2019video,Yang_2021_ICCV} have largely been replaced by end-to-end transformer frameworks. Offline methods such as SeqFormer~\cite{wu2022seqformer} process an entire clip at once through video-level object queries, whereas online methods such as GenVIS~\cite{heo2023generalized}, TCOVIS~\cite{li2023tcovis}, DVIS~\cite{zhang2023dvis}, and VideoMT~\cite{norouzi2026videomt} propagate instance queries frame by frame. Either way, both paradigms require multi-frame annotations with consistent identities across the clip. MinVIS~\cite{huang2022minvis} is the main exception and the line closest to ours: it trains on frames as independent images and associates instances only at inference, showing that a strong image model is already a competitive VIS baseline without video supervision. Its queries, however, are optimized by generic classification and mask losses and drift apart under occlusion and appearance change, which destabilizes association. QueenVIS keeps this image-only regime but closes much of the gap by enriching queries with appearance and spatial objectives during single-frame training.
	
	\subsection{Cross-Frame Association and Tracking}
	Beyond the modeling paradigm, VIS methods differ in how they associate instances across frames, and this association is almost always a learned module. Online trackers match per-frame predictions through contrastive embeddings, as in IDOL~\cite{wu2022defense}, or propagate dedicated track queries~\cite{meinhardt2023novis,wu2022efficient}. A common ingredient is an explicit memory bank of instance features that resolves long occlusions, used by CTVIS~\cite{ying2023ctvis}, VISOLO~\cite{han2022visolo}, and VISAGE~\cite{kim2024visage}. In CTVIS and VISOLO the memory is written and read by video-trained modules; VISAGE instead employs a non-parametric, training-free memory bank, though it stores both object queries and dedicated appearance features produced by a video-supervised network. QueenVIS keeps association entirely training-free \emph{and} feeds it from an image-only model: confidence-guided blending warm-starts the current frame with high-confidence queries from previous frames, paired with the non-parametric memory bank of VISAGE~\cite{kim2024visage} repurposed to store our object queries directly for bipartite matching, injecting temporal stability at inference without any tracking parameters or video supervision.
	
	\begin{figure*}[!t]
		\centering
		\includegraphics[width=1.0\linewidth]{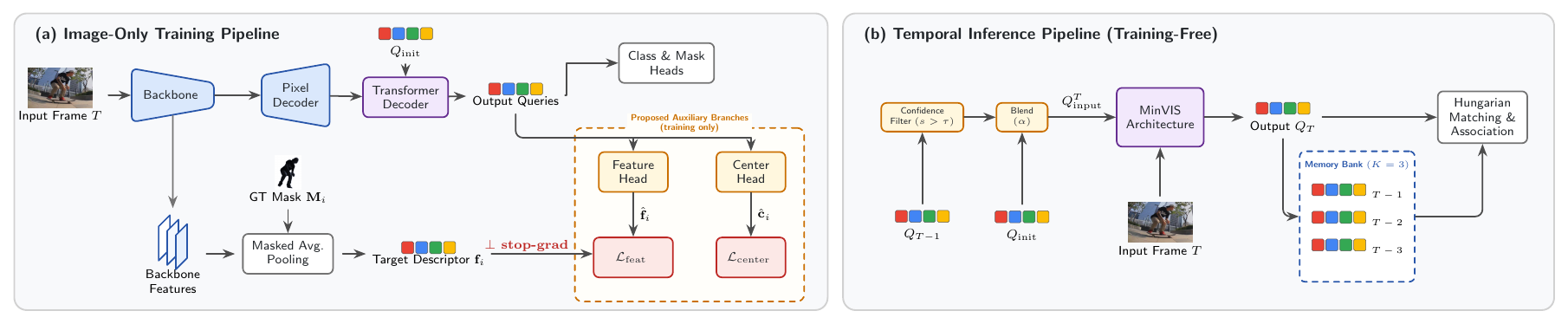}
		\caption{\textbf{Overview of the proposed appearance-aware query learning framework.} To preserve the efficiency of image-only optimization, the architecture is divided into two distinct phases. \textbf{(a) Image-Only Training Pipeline:} We introduce auxiliary feature and center prediction branches (active only during training). Ground-truth features are extracted by pooling backbone feature maps using target masks, supervising the transformer queries via an $\ell_1$ loss to enforce dense appearance awareness. \textbf{(b) Temporal Inference Pipeline:} During inference, the auxiliary branches are discarded. Temporal identity is maintained entirely without video-level training using a training-free query propagation mechanism, which blends high-confidence queries from the previous frame ($T-1$) with the current frame alongside a lightweight memory bank ($K=3$) for robust historical matching.}
		\label{fig:main_architecture}
	\end{figure*}
	
	\section{Method}
	
	\subsection{Overview}
	QueenVIS (Figure~\ref{fig:main_architecture}) trains a Mask2Former-based architecture under strictly single-frame supervision. During training, two auxiliary heads enrich each query: a feature-prediction head that distills the backbone's per-instance appearance descriptor into the query by regressing the query toward that descriptor (Section~\ref{sec:feat}), and a center-prediction head that injects a spatial prior (Section~\ref{sec:center}). Both heads are discarded afterward, adding no inference cost. At inference, temporal identity is maintained without any learned tracker through confidence-guided query propagation and a non-parametric memory bank (Section~\ref{sec:assoc}).
	
	\subsection{Appearance-Aware Query Learning}
	\label{sec:feat}
	As argued in Section~\ref{sec:intro}, the appearance signal that video-supervised methods learn from clips is already latent in per-frame features; here we make it explicit and inject it into the object query under image-only supervision. Standard Mask2Former queries are optimized only by classification and mask losses. The backbone, however, already computes a discriminative per-instance appearance signal, yet nothing in the standard objective requires the object query to retain it. We therefore treat the backbone's pooled instance descriptor as a fixed teacher target and distill it into the corresponding query: an auxiliary head projects each query into the descriptor space, and an L1 loss forces the query to reproduce the backbone signal it would otherwise drop.
	
	Given backbone feature maps ${\mathbf{F}^{(l)}} \in \mathbb{R}^{C_l \times H_l \times W_l}$ and a ground-truth object mask $\mathbf{M}_i$, we extract object-level appearance descriptors via masked average pooling:
	\begin{equation}
		\mathbf{f}_i^{(l)} = 
		\frac{
			\sum_{u,v}
			\tilde{\mathbf{M}}_i^{(l)}(u,v) \odot \mathbf{F}^{(l)}(:,u,v)
		}{
			\sum_{u,v}
			\tilde{\mathbf{M}}_i^{(l)}(u,v)
		},
	\end{equation}
	where $\tilde{\mathbf{M}}_i^{(l)}$ is the resized mask. The descriptors from the 4th and 5th backbone stages are concatenated to form the final target representation $\mathbf{f}_i = [\mathbf{f}_i^{(4)}, \mathbf{f}_i^{(5)}]$.
	
	An auxiliary MLP head maps each object query to a prediction of its corresponding descriptor. Because object queries are unordered, the regression is defined on matched query--instance pairs. Let $\sigma$ denote Mask2Former's Hungarian assignment, mapping ground-truth instance $i$ to the predicted query slot $\sigma(i)$. We $\ell_2$-normalize both vectors and apply an $\ell_1$ regression loss:
	\begin{equation}
		\mathcal{L}_{\text{feat}}=
		\frac{1}{N}
		\sum_{i=1}^{N}
		\left\|
		\frac{\hat{\mathbf{f}}_{\sigma(i)}}
		{\|\hat{\mathbf{f}}_{\sigma(i)}\|_2} -
		\frac{\mathbf{f}_i}
		{\|\mathbf{f}_i\|_2}
		\right\|_1
	\end{equation}
	
	As supported by our diagnostic analysis (Section~\ref{sec:diagnostic}, Table~\ref{tab:oracle_analysis}), enriching the object queries helps relieve the temporal-association bottleneck: QueenVIS cuts the T-Oracle tracking gap from $12.5$ (MinVIS) to $7.5$ AP, making cross-frame association more stable and discriminative.
	
	\subsection{Center Prediction}
	\label{sec:center}
	A prevalent failure mode in online VIS is the long-range identity switch, where after heavy occlusion a track jumps to a different instance in a distant region of the frame. Appearance cannot resolve these cases when the competing instances share a category and look alike, as in crowded OVIS scenes where location is the only separating cue; center prediction therefore complements the feature objective of Section~\ref{sec:feat} along an orthogonal, spatial axis. Standard Mask2Former queries carry some latent position sensitivity~\cite{kim2024visage}, which we make explicit through direct supervision; since spatial coordinates are simple to regress, a single linear projection, discarded at inference with no added parameters, suffices and yields a consistent $+1.3$ AP on YouTube-VIS 2021 (Table~\ref{tab:ablation_components}). Let $\mathbf{c}_i^* \in [0, 1]^2$ denote the normalized center coordinates of the bounding box for the ground-truth mask $\mathbf{M}_i$. We attach a linear projection layer to each object query to predict its normalized spatial coordinates $\hat{\mathbf{c}} = \text{Linear}(\mathbf{Q})$. Utilizing the same Hungarian assignment $\sigma$, the center prediction objective is supervised via an $\ell_1$ loss:
	\begin{equation}
		\mathcal{L}_{\text{center}} = \frac{1}{N} \sum_{i=1}^{N} \left\| \hat{\mathbf{c}}_{\sigma(i)} - \mathbf{c}_i^* \right\|_1
	\end{equation}
	Regressing each query to its absolute center gives the bipartite matcher an explicit geometric prior that stabilizes association and reduces identity switches across distant spatial locations.

	\subsection{Training Objective}
	The two auxiliary objectives are optimized jointly with the standard baseline losses under single-frame supervision. The overall objective adds the feature- and center-prediction terms to the MinVIS losses ($\lambda_{\text{feat}}=0.2$, $\lambda_{\text{center}}=0.05$):
	\begin{equation}
		\mathcal{L} = \mathcal{L}_{\text{MinVIS}} + \lambda_{\text{feat}}\mathcal{L}_{\text{feat}} + \lambda_{\text{center}}\mathcal{L}_{\text{center}}
	\end{equation}
	where $\mathcal{L}_{\text{MinVIS}} = \lambda_{\text{ce}}\mathcal{L}_{\text{ce}} + \lambda_{\text{dice}}\mathcal{L}_{\text{dice}} + \lambda_{\text{mask}}\mathcal{L}_{\text{mask}}$ denotes the baseline objective ($\lambda_{\text{ce}}=2.0$, $\lambda_{\text{dice}}=\lambda_{\text{mask}}=5.0$). Beyond serving as a loss, the feature-prediction distance also augments the bipartite matching cost: extending the Mask2Former matching cost~\cite{cheng2021mask2former}, the Hungarian algorithm that assigns ground-truth instances to query slots within each frame minimizes the classification and mask costs together with the $\ell_1$ appearance distance, so that appearance agreement guides the assignment $\sigma$ itself rather than only its subsequent supervision. Both auxiliary heads are removed after training, so the segmentation network carries no additional parameters over MinVIS; temporal identity is instead maintained by the training-free association scheme described next.

	\subsection{Training-Free Temporal Association}
	\label{sec:assoc}
	To inject temporal consistency at inference without a learned tracker, we blend queries across frames under confidence guidance. Object queries decouple into learnable content queries $\mathbf{C}_i$ and static positional queries $\mathbf{P}_i$; to preserve the native spatial prior, we propagate only the content embeddings and leave $\mathbf{P}_i$ unchanged. Let $\mathbf{C}_{i,\text{init}}^{t}$ be the default initialization of content query $i$ at frame $t$. If its previous-frame score $s_i^{t-1}$ exceeds a threshold $\tau$, we blend the optimized embedding $\mathbf{C}_i^{t-1}$ into the same query at the current frame:
	\begin{equation}
		\mathbf{C}_i^{t} =
		\begin{cases}
			\alpha \mathbf{C}_i^{t-1} + (1-\alpha) \mathbf{C}_{i,\text{init}}^{t} & \text{if } s_i^{t-1} > \tau \\
			\mathbf{C}_{i,\text{init}}^{t} & \text{otherwise}
		\end{cases}
	\end{equation}
	The blending factor $\alpha$ acts as a regularizer: it injects a temporal ``warm start'' from the previous frame while keeping the input within the decoder's expected manifold, mitigating train--test distribution shift. Low-confidence queries, e.g.\ under severe occlusion, fall back to the learned prior $\mathbf{C}_{i,\text{init}}^{t}$, and the decoder corrects any remaining drift.

	Propagation relies on a single previous frame, so a momentary drop in quality, e.g.\ a brief occlusion, can destabilize association. To make matching more robust, we keep a lightweight, non-parametric memory bank that aggregates each instance's embeddings over the most recent $K=3$ frames into one temporal embedding by confidence-aware temporal weighting, adapted from VISAGE~\cite{kim2024visage}. As in query-based VIS, identities are then assigned by Hungarian matching between the current-frame queries and these stored embeddings using embedding similarity; matched queries inherit the existing identity, and unmatched high-confidence queries start new tracks.

	\begin{table*}[!t]
		\centering
		\caption{Comparison with state-of-the-art VIS methods. Metrics denote Average Precision (AP) at various IoU thresholds and Average Recall (AR). Methods are grouped by their reliance on explicit video supervision and by the online/offline inference setting. Bold numbers denote the best result among image-only methods.}
		\label{tab:main_comparison}
		\resizebox{\linewidth}{!}{
			\begin{tabular}{l c | ccccc | ccccc | ccccc}
				\toprule
				\multirow{2}{*}{Method} & \multirow{2}{*}{Setting} & \multicolumn{5}{c|}{OVIS} & \multicolumn{5}{c|}{YouTube-VIS 2019} & \multicolumn{5}{c}{YouTube-VIS 2021} \\
				\cmidrule(lr){3-7} \cmidrule(lr){8-12} \cmidrule(lr){13-17}
				& & AP & AP$_{50}$ & AP$_{75}$ & AR$_1$ & AR$_{10}$ & AP & AP$_{50}$ & AP$_{75}$ & AR$_1$ & AR$_{10}$ & AP & AP$_{50}$ & AP$_{75}$ & AR$_1$ & AR$_{10}$ \\
				\midrule
				\multicolumn{17}{c}{\textbf{ResNet-50 Backbone}}\\
				\midrule
				\multicolumn{17}{l}{\textit{Offline Methods (Video Supervision)}}\\
				\midrule
				IFC~\cite{hwang2021video} & Offline & 13.1 & 27.8 & 11.6 & 9.4 & 23.9 & 41.2 & 65.1 & 44.6 & 42.3 & 49.6 & 35.2 & 55.9 & 37.7 & 32.6 & 42.9 \\
				SeqFormer~\cite{wu2022seqformer} & Offline & 15.1 & 31.9 & 13.8 & 10.4 & 27.1 & 47.4 & 69.8 & 51.8 & 45.5 & 54.8 & 40.5 & 62.4 & 43.7 & 36.1 & 48.1 \\
				Mask2Former-VIS~\cite{cheng2021mask2former} & Offline & 17.3 & 37.3 & 15.1 & 10.5 & 23.5 & 46.4 & 68.0 & 50.0 & -- & -- & 40.6 & 60.9 & 41.8 & -- & -- \\
				VITA~\cite{heo22vita} & Offline & 19.6 & 41.2 & 17.4 & 11.7 & 26.0 & 49.8 & 72.6 & 54.5 & 49.4 & 61.0 & 45.7 & 67.4 & 49.5 & 40.9 & 53.6 \\
				MDQE~\cite{li2023mdqe} & Offline & 29.2 & 55.2 & 27.1 & 14.5 & 34.2 & 47.8 & -- & -- & -- & -- & 44.5 & 67.1 & 48.7 & 37.9 & 49.8 \\
				NOVIS~\cite{meinhardt2023novis} & Offline & 30.8 & 54.4 & 31.0 & 15.2 & 35.3 & 52.8 & 75.7 & 56.9 & 50.3 & 60.6 & 47.2 & 69.4 & 50.0 & 41.3 & 54.4 \\
				RefineVIS~\cite{abrantes2023refinevis} & Offline & 33.7 & 56.2 & 34.8 & 15.6 & 39.8 & 52.2 & 76.3 & 57.7 & 47.5 & 57.6 & 50.2 & 72.8 & 55.4 & 41.2 & 56.3 \\
				DVIS~\cite{zhang2023dvis} & Offline & 33.8 & 60.4 & 33.5 & 15.3 & 39.5 & 52.6 & 76.5 & 58.2 & 47.4 & 60.4 & 47.4 & 71.0 & 51.6 & 39.9 & 55.2 \\
				GenVIS~\cite{heo2023generalized} & Offline & 34.5 & 59.4 & 35.0 & 16.6 & 38.3 & 51.3 & 72.0 & 57.8 & 49.5 & 60.0 & 46.3 & 67.0 & 50.2 & 40.6 & 53.2 \\
				\midrule
				\multicolumn{17}{l}{\textit{Online Methods (Video Supervision)}}\\
				\midrule
				MaskTrack R-CNN~\cite{yang2019video} & Online & 10.8 & 25.3 & 8.5 & 7.9 & 14.9 & 30.3 & 51.1 & 32.6 & 31.0 & 35.5 & 28.6 & 48.9 & 29.6 & 26.5 & 33.8 \\
				SipMask~\cite{cao2020sipmask} & Online & 10.2 & 24.7 & 7.8 & 7.9 & 15.8 & 33.7 & 54.1 & 35.8 & 35.4 & 40.1 & 31.7 & 52.5 & 34.0 & 30.8 & 37.8 \\
				CrossVIS~\cite{Yang_2021_ICCV} & Online & 14.9 & 32.7 & 12.1 & 10.3 & 19.8 & 36.3 & 56.8 & 38.9 & 35.6 & 40.7 & 34.2 & 54.4 & 37.9 & 30.4 & 38.2 \\
				VISOLO~\cite{han2022visolo} & Online & 15.3 & 31.0 & 13.8 & 11.1 & 21.7 & 38.6 & 56.3 & 43.7 & 35.7 & 42.5 & 36.9 & 54.7 & 40.2 & 30.6 & 40.9 \\
				IDOL~\cite{wu2022defense} & Online & 28.2 & 51.0 & 28.0 & 14.5 & 38.6 & 49.5 & 74.0 & 52.9 & 47.7 & 58.7 & 43.9 & 68.0 & 49.6 & 38.0 & 50.9 \\
				DVIS~\cite{zhang2023dvis} & Online & 30.2 & 55.0 & 30.5 & 14.5 & 37.3 & 51.2 & 73.8 & 57.1 & 47.2 & 59.3 & 46.4 & 68.4 & 49.6 & 39.7 & 53.5 \\
				TCOVIS~\cite{li2023tcovis} & Online & 35.3 & 60.7 & 36.6 & 15.7 & 39.5 & 52.3 & 73.5 & 57.6 & 49.8 & 60.2 & 49.5 & 71.2 & 53.8 & 41.3 & 55.9 \\
				CTVIS~\cite{ying2023ctvis} & Online & 35.5 & 60.8 & 34.9 & 16.1 & 41.9 & 55.1 & 78.2 & 59.1 & 51.9 & 63.2 & 50.1 & 73.7 & 54.7 & 41.8 & 59.5 \\
				GenVIS~\cite{heo2023generalized} & Online & 35.8 & 60.8 & 36.2 & 16.3 & 39.6 & 50.0 & 71.5 & 54.6 & 49.5 & 59.7 & 47.1 & 67.5 & 51.5 & 41.6 & 54.7 \\
				VISAGE~\cite{kim2024visage} & Online & 36.2 & 60.3 & 35.3 & 17.0 & 40.3 & 55.1 & 78.1 & 60.6 & 51.0 & 62.3 & 51.6 & 73.8 & 56.1 & 43.6 & 59.3 \\
				CAVIS~\cite{lee2025cavis} & Online & 37.6 & 63.4 & 38.2 & 16.5 & 43.5 & 55.7 & 78.3 & 61.7 & 51.5 & 63.3 & 50.5 & 74.1 & 54.9 & 42.6 & 58.5 \\
				\midrule
				\multicolumn{17}{l}{\textit{Image-Only Training (No Video Supervision)}}\\
				\midrule
				MinVIS~\cite{huang2022minvis} & Online & 25.0 & 45.5 & 24.0 & 13.9 & 29.7 & 47.4 & 69.0 & 52.1 & 45.7 & 55.7 & 44.2 & 66.0 & 48.1 & 39.2 & 51.7 \\
				\textbf{QueenVIS} & Online & \textbf{29.8} & \textbf{52.7} & \textbf{29.0} & \textbf{15.4} & \textbf{34.6} & \textbf{51.8} & \textbf{75.2} & \textbf{58.3} & \textbf{49.4} & \textbf{62.1} & \textbf{50.9} & \textbf{72.1} & \textbf{55.8} & \textbf{42.8} & \textbf{57.3} \\
				\bottomrule
			\end{tabular}
		}
	\end{table*}
	
	\section{Experiments}
	
	\subsection{Datasets \& Implementation Details}
	We evaluate QueenVIS on four standard VIS benchmarks: YouTube-VIS 2019 and YouTube-VIS 2021~\cite{yang2019video,vis2021}, the YouTube-VIS 2022 \emph{long} split~\cite{vis2022}, whose substantially longer sequences stress long-horizon temporal association, and the more challenging, heavily occluded OVIS~\cite{qi2022ovis}. We report the standard video Average Precision (AP) and Average Recall (AR) metrics. All experiments use a ResNet-50~\cite{he2016resnet} backbone. We follow the VISAGE~\cite{kim2024visage} training recipe unchanged except that we remove its video-level supervision and train on single frames only, adding our two auxiliary objectives so that any gain is attributable to query enrichment rather than to retuning the baseline. Models are trained on a single NVIDIA H200 GPU with \texttt{num\_frames=1} (a single frame per training step): 75k iterations for OVIS and YouTube-VIS 2021 and 30k for YouTube-VIS 2019. YouTube-VIS 2022 shares its training data with YouTube-VIS 2021, so we evaluate the YouTube-VIS 2021 model directly on the 2022 \emph{long} validation set without any additional training. At inference, the training-free association uses a memory bank of size $K=3$ and confidence-guided query propagation with blending factor $\alpha=0.25$ and confidence threshold $\tau=0.8$.
	
	\begin{table*}[t]
		\centering
		\caption{Incremental ablation study on the YouTube-VIS 2021 validation set. QP: Query Propagation, MB: Memory Bank, FP: Feature Prediction loss, CP: Center Prediction loss.}
		\label{tab:ablation_components}
		\begin{tabular}{l l c c c}
			\toprule
			Method & Configuration & AP & AP$_{50}$ & AP$_{75}$ \\
			\midrule
			MinVIS~\cite{huang2022minvis} & Official model & 44.2 & 66.0 & 48.1 \\
						MinVIS & Official + QP + MB & 46.4 & 70.1 & 50.1 \\
			QueenVIS & FP ($\lambda_{\text{feat}}=0.20$) + QP + MB & 49.6 & 71.1 & 54.7 \\
			QueenVIS & FP ($0.20$) + CP ($0.05$) + QP + MB & \textbf{50.9} & \textbf{72.1} & \textbf{55.8} \\
			\bottomrule
		\end{tabular}
	\end{table*}	
	\subsection{Main Results}
	Table~\ref{tab:main_comparison} compares QueenVIS against state-of-the-art VIS methods on OVIS, YouTube-VIS 2019, and YouTube-VIS 2021. Across all three benchmarks, enriching object queries during single-frame training consistently improves the image-only baseline. QueenVIS raises MinVIS from 44.2 to 50.9 AP on YouTube-VIS 2021 (+6.7 AP) and from 47.4 to 51.8 AP on YouTube-VIS 2019 (+4.4 AP). The improvement is also substantial on OVIS, where occlusion is most severe and appearance- and spatial-aware queries matter most: QueenVIS improves from 25.0 to 29.8 AP (+4.8 AP), a nearly +20\% relative gain over the baseline. None of these gains require video clips, temporal gradients, or any added inference parameters.

	Despite relying solely on image-level supervision, QueenVIS substantially narrows the gap to recent video-supervised methods. On YouTube-VIS 2021 its 50.9 AP surpasses the online video-supervised CAVIS~\cite{lee2025cavis} (50.5 AP) and trails the strongest video-supervised method, VISAGE~\cite{kim2024visage} (51.6 AP), by only 0.7 AP, bringing image-only training to within run-to-run variation of video supervision on this benchmark. On YouTube-VIS 2019, QueenVIS (51.8 AP) surpasses IDOL~\cite{wu2022defense} and GenVIS~\cite{heo2023generalized} while trailing the strongest video-supervised methods by a modest margin. The picture is different on OVIS: QueenVIS (29.8 AP) improves over IDOL~\cite{wu2022defense} (28.2 AP), but a substantial gap remains to most video-supervised methods in Table~\ref{tab:main_comparison}, including NOVIS, TCOVIS, CTVIS, VISAGE, and CAVIS. OVIS represents the most extreme occlusion regime, where the limit of image-only training is most visible; we argue this gap is a target for better-designed auxiliary objectives rather than an inherent ceiling. Query enrichment already recovers most of the gain previously credited to video supervision on the standard benchmarks, and the remaining margin under dense occlusion motivates further research in this direction.

	\textbf{Long-video setting.} We further evaluate on the YouTube-VIS 2022 \emph{long} validation set, whose sequences are markedly longer than those in the 2019/2021 splits and therefore stress long-horizon temporal association, exactly the regime in which per-frame query drift accumulates most. Table~\ref{tab:ytvis22_long} reports the comparison against the MinVIS baseline under the same ResNet-50 backbone. QueenVIS improves AP from $23.3$ to $\mathbf{33.6}$ (\textbf{+10.3 AP}), a larger margin than on any of the shorter benchmarks, with consistent gains across all metrics (e.g.\ +14.7 AP$_{75}$ and +11.5 AR$_1$). This widening gap on long videos indicates that appearance- and spatial-aware queries become increasingly valuable as sequence length grows, supporting our claim that query quality, rather than video-level supervision, is the binding constraint on temporal association.

	\begin{table}[tb]
		\centering
		\caption{Comparison on the YouTube-VIS 2022 \emph{long} validation set (ResNet-50 backbone), whose long sequences emphasize long-horizon association. QueenVIS improves over the image-only MinVIS baseline without any video supervision.}
		\label{tab:ytvis22_long}
		\resizebox{\linewidth}{!}{
			\begin{tabular}{l ccccc}
				\toprule
				Method & AP & AP$_{50}$ & AP$_{75}$ & AR$_1$ & AR$_{10}$ \\
				\midrule
				MinVIS~\cite{huang2022minvis} & 23.3 & 47.9 & 19.3 & 20.2 & 28.0 \\
				\textbf{QueenVIS} & \textbf{33.6} & \textbf{56.6} & \textbf{34.0} & \textbf{31.7} & \textbf{39.9} \\
				\bottomrule
			\end{tabular}
		}
	\end{table}
	
	\begin{figure*}[!t]
		\centering
		\includegraphics[width=0.75\linewidth]{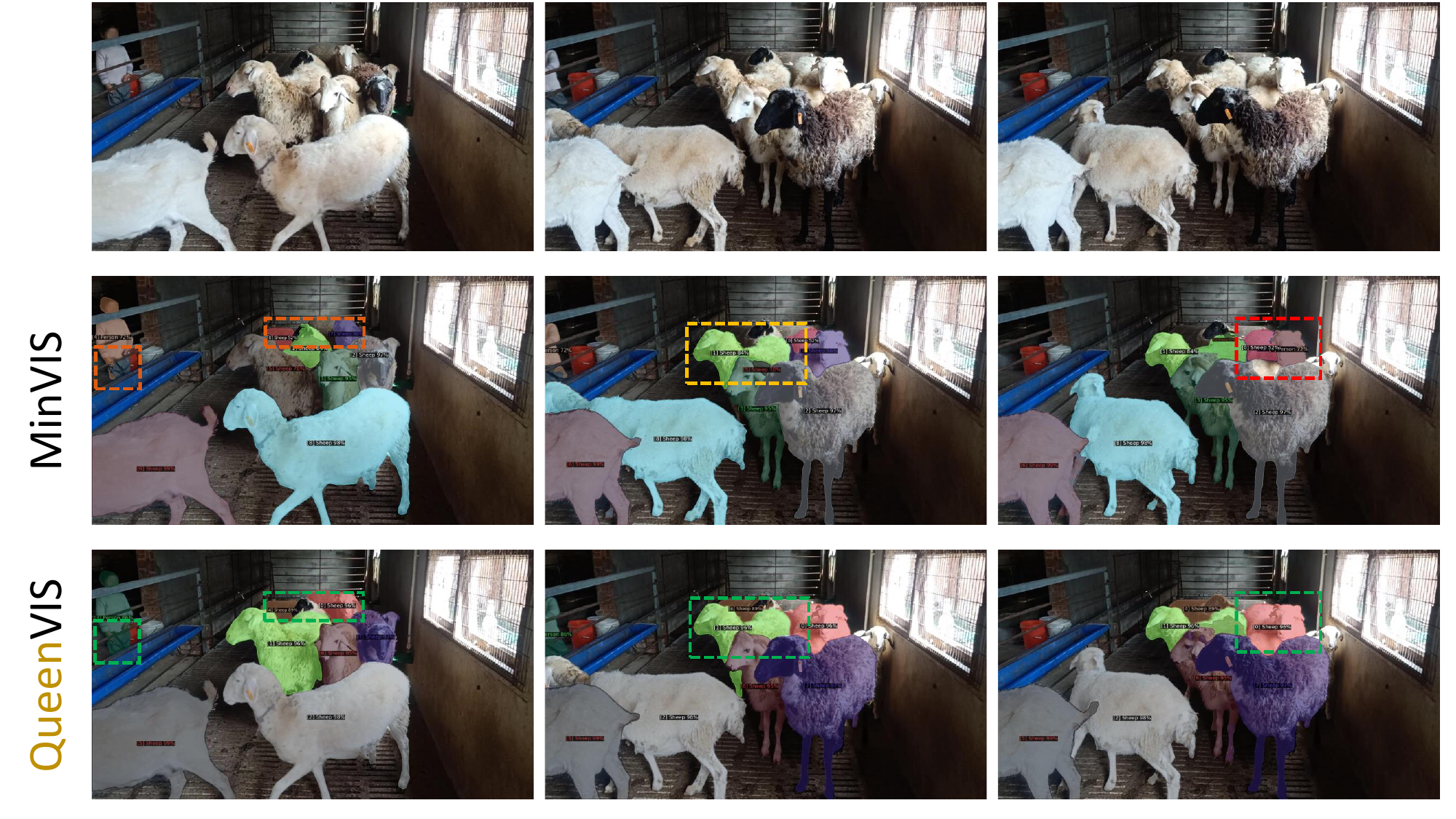}
		\caption{ Qualitative comparison on the OVIS dataset. The rows correspond to the original input frames (top), MinVIS baseline predictions (middle), and our proposed QueenVIS predictions (bottom). Left: MinVIS exhibits mask fragmentation and poor segmentation quality (orange dashed rectangles), which QueenVIS resolves to produce coherent, accurate masks (green dashed rectangles). Middle: During heavy occlusion, MinVIS suffers from a severe identity switch, confusing two different instances (yellow dashed rectangle). QueenVIS successfully maintains a consistent temporal track for the correct identity. Right: When a target (the person) disappears from the scene, MinVIS incorrectly assigns the obsolete track to a sheep's face (red dashed rectangle). By effectively enriching queries, QueenVIS accurately handles the disappearance and suppresses false positive assignments.}
		\label{fig:qualitative}
	\end{figure*}	
	
	\subsection{Qualitative Results}
	Figure~\ref{fig:qualitative} compares QueenVIS against MinVIS on OVIS sequences with heavy occlusion. When trajectories intersect, the baseline matches queries that lack discriminative appearance cues and frequently fails, producing identity switches or fragmented masks. QueenVIS instead preserves identities through the occlusion: by embedding dense appearance and spatial priors into the object queries during training, it re-identifies each target after it reappears, without any temporal video supervision. As a side benefit, these same enriched queries also yield more complete, less fragmented masks. These cases illustrate the same failure mode that the diagnostic analysis below quantifies.
	
	\subsection{Diagnostic Bottleneck Analysis}
	\label{sec:diagnostic}
	A raw AP gain is ambiguous: it may stem from sharper masks or more accurate classification rather than from better tracking, so AP alone cannot tell us whether query enrichment actually relieves the \emph{association} bottleneck of online VIS or merely improves these other factors. To attribute the gain to a specific error source, we evaluate QueenVIS with the model-agnostic oracle framework of Hamdi et al.~\cite{hamdi2026mind}. The framework replaces one error source at a time with a ground-truth oracle: a Tracking Oracle (T-Oracle) isolates pure temporal-association error, while a Tracking+Classification Oracle (TC-Oracle) additionally removes semantic-assignment error, leaving only mask-representation quality. The reported gap $\Delta$ is the AP a model leaves on the table relative to the corresponding oracle, so a smaller $\Delta$ means the model operates closer to its own upper bound. Table~\ref{tab:oracle_analysis} reports both gaps for QueenVIS and MinVIS on the YouTube-VIS 2021 validation set.
	
	The two gaps reveal distinct behavior in tracking stability and in classification robustness, which we examine in turn.
	
	\begin{table}[tb]
		\centering
		\caption{Diagnostic error gap analysis on YouTube-VIS 2021 using the framework of Hamdi et al.~\cite{hamdi2026mind}. Lower values ($\Delta$) indicate a tighter performance gap to the respective optimized global oracles. VISAGE and CAVIS are video-supervised; QueenVIS attains the smallest (or tied-smallest) gaps despite image-only training.}
		\label{tab:oracle_analysis}
		\resizebox{\linewidth}{!}{
			\begin{tabular}{l cc cc}
				\toprule
				\multirow{2}{*}{Method} & \multicolumn{2}{c}{T-Oracle Gap ($\Delta_{TO}$)} & \multicolumn{2}{c}{TC-Oracle Gap ($\Delta_{TCO}$)} \\
				\cmidrule(lr){2-3} \cmidrule(lr){4-5}
				& AP & $\text{AR}_1$ & AP & $\text{AR}_1$ \\
				\midrule
				MinVIS~\cite{huang2022minvis} & 12.5 & 7.9 & 20.0 & 11.4 \\
				VISAGE~\cite{kim2024visage} & 9.2 & 6.2 & 16.7 & 9.8 \\
				CAVIS~\cite{lee2025cavis} & 8.7 & 4.2 & \textbf{14.2} & 8.5 \\
				\textbf{QueenVIS} & \textbf{7.5} & \textbf{1.0} & \textbf{14.2} & \textbf{5.4} \\
				\bottomrule
			\end{tabular}
		}
	\end{table}
	
	\textbf{Tracking robustness.} The T-Oracle gap measures error attributable to association alone, so it serves as a direct proxy for identity consistency that overall AP hides. MinVIS leaves a gap of $12.5$ AP and $7.9$ $\text{AR}_1$. QueenVIS cuts the tracking gap to $\mathbf{7.5}$ AP and, most strikingly, the tracking-recall gap to $\mathbf{1.0}$ $\text{AR}_1$, essentially the oracle ceiling, so almost every instance it detects is also tracked correctly. This $7.5$ AP tracking gap is even tighter than that of strong video-supervised trackers such as VISAGE ($9.2$ AP)~\cite{kim2024visage} and CAVIS ($8.7$ AP)~\cite{lee2025cavis}, despite QueenVIS never seeing a video clip: the embedded appearance and spatial priors and inference-time history close most of the association gap once thought to require video-level temporal gradients.
	
	\textbf{Semantic and combined errors.} Relaxing the classification constraint with the TC-Oracle exposes the combined classification-and-matching headroom. QueenVIS reduces this gap from $20.0$ AP (MinVIS) to $\mathbf{14.2}$ AP. Subtracting the tracking gap isolates the classification-only component ($\Delta_{TCO}-\Delta_{TO}$): it is $7.5$ AP for MinVIS and $6.7$ AP for QueenVIS, essentially unchanged. Almost all of our improvement therefore comes from tracking, and the residual headroom is now split evenly between association and semantics rather than dominated by tracking; the diagnostic itself singles out the classification head as the next lever. Even so, this $14.2$ AP combined gap ranks among the tightest reported for recent online VIS systems by Hamdi et al.~\cite{hamdi2026mind}, matching video-supervised CAVIS ($14.2$ AP)~\cite{lee2025cavis} and undercutting VISAGE ($16.7$ AP)~\cite{kim2024visage} even though QueenVIS trails it by only $0.7$ AP in raw accuracy: QueenVIS reaches a comparable score while operating markedly closer to its own upper bound.
	
	\subsection{Ablation Studies}
	Table~\ref{tab:ablation_components} isolates each component on YouTube-VIS 2021. Applied to the official MinVIS weights, training-free query propagation and the memory bank lift AP from $44.2$ to $46.4$ (+2.2) without any retraining. The larger improvement, however, comes from training rather than inference: making the queries appearance- and spatial-aware through feature and center prediction adds a further +4.5 AP, reaching $50.9$. This ordering supports our central claim that query representation, not the association heuristic, is the binding constraint in image-only VIS.
	
	We also considered a relational alternative to our per-query feature loss, distilling the pairwise similarity geometry between queries rather than their absolute descriptors~\cite{park2019rkd}. In our experiments this consistently degraded performance; we report the full formulation and ablation in the supplementary material, along with a sensitivity analysis of the loss weights $\lambda_{\text{feat}}$ and $\lambda_{\text{center}}$.

	\textbf{Which feature level to distill.} The feature-prediction target can be drawn from different backbone stages, and this choice matters. Table~\ref{tab:ablation_features} compares descriptors pooled from the intermediate stage (Res4), the deepest stage (Res5), and their concatenation. Res5 carries strong semantics but coarse localization, while Res4 carries finer spatial detail but weaker semantics; concatenating the two (Res4 + Res5) performs best, supplying the query with both the semantic and the spatial granularity that cross-frame association needs.
	
	\begin{table}[tb]
		\centering
		\caption{Effect of target feature levels for appearance prediction on YouTube-VIS 2021 (ResNet-50). Concatenating intermediate and deep features yields the best representation.}
		\label{tab:ablation_features}
		\begin{tabular}{lccc}
			\toprule
			Feature Level & AP & AP$_{50}$ & AP$_{75}$ \\
			\midrule
			Res4 & 48.7 & 70.0 & 52.5 \\
			Res5 & 49.2 & 70.6 & 54.2 \\
			Res4 + Res5 (QueenVIS) & \textbf{50.9} & \textbf{72.1} & \textbf{55.8} \\
			\bottomrule
		\end{tabular}
	\end{table}
	
	\textbf{Inference hyperparameters.} The association pipeline exposes three inference-time hyperparameters: the memory-bank size $K$, the query-propagation blending factor $\alpha$, and the propagation confidence threshold $\tau$. Table~\ref{tab:ablation_hyperparams} reports a sensitivity sweep over each on YouTube-VIS 2021, varying one at a time around the default $(K{=}3,\,\alpha{=}0.25,\,\tau{=}0.80)$. The variance across configurations is negligible: the gain stems from the presence of query propagation rather than its precise tuning. We keep this default unchanged across all datasets, including the heavily occluded OVIS.
	
	\begin{table}[tb]
		\centering
		\caption{Ablation on Query Propagation and Memory Bank hyperparameters on YouTube-VIS 2021 (ResNet-50). We evaluate memory bank size ($K$), blending factor ($\alpha$), and confidence threshold ($\tau$). Default settings are highlighted in bold.}
		\label{tab:ablation_hyperparams}
		\begin{tabular}{ccc|ccc}
			\toprule
			$K$ & $\alpha$ & $\tau$ & AP & AP$_{50}$ & AP$_{75}$ \\
			\midrule
			1 & 0.25 & 0.80 & 49.3 & 70.2 & 53.7 \\
			\textbf{3} & \textbf{0.25} & \textbf{0.80} & \textbf{50.9} & \textbf{72.1} & \textbf{55.8} \\
			5 & 0.25 & 0.80 & 50.5 & 71.6 & 55.7 \\
			\midrule
			3 & 0.1 & 0.80 & 50.4 & 71.3 & 55.0 \\
			\textbf{3} & \textbf{0.25} & \textbf{0.80} & \textbf{50.9} & \textbf{72.1} & \textbf{55.8} \\
			3 & 0.5 & 0.80 & 50.4 & 71.2 & 55.8 \\
			\midrule
			3 & 0.25 & 0.50 & 50.4 & 71.7 & 55.2 \\
			\textbf{3} & \textbf{0.25} & \textbf{0.80} & \textbf{50.9} & \textbf{72.1} & \textbf{55.8} \\
			3 & 0.25 & 0.90 & 50.4 & 71.4 & 55.6 \\
			\bottomrule
		\end{tabular}
	\end{table}	
	\textbf{Inference efficiency.} QueenVIS adds zero learnable parameters at inference and maintains the exact same GFLOPs as the MinVIS baseline. The training-free memory bank incurs only a marginal computational overhead, resulting in a throughput drop of less than 7 FPS. This trades a negligible, comfortably real-time cost for substantial accuracy gains without the massive scaling penalties of video-supervised trackers.
	\section{Conclusion}
	This work explores the capabilities of image-only training for competitive VIS. By restructuring the representation learning of transformer queries during single-frame optimization, we force the network to learn highly discriminative, instance-specific appearance and geometry priors. When paired with a training-free query propagation strategy, QueenVIS successfully maintains temporal identity without processing video clips. Consequently, our approach bridges the performance gap, rivaling state-of-the-art video-supervised architectures while significantly reducing training complexity. The benefit is largest on longer videos, where QueenVIS outperforms MinVIS by $+10.3$ AP on the long-sequence YouTube-VIS 2022 split. On YouTube-VIS 2021, the model-agnostic oracle framework of Hamdi et al.~\cite{hamdi2026mind} (T-Oracle and TC-Oracle) attributes QueenVIS's gains almost entirely to improved association, narrowing the tracking-recall gap to within $1.0$ $\text{AR}_1$ of the oracle ceiling, tighter than even video-supervised trackers. A performance gap nonetheless persists under the densest occlusions of OVIS, leaving a detailed diagnosis of these failure cases and targeted improvements to future work. Our results suggest that strong spatial and appearance modeling can substantially reduce the performance gap associated with the absence of explicit temporal supervision.
	
	{\small
		\bibliographystyle{ieeetr}
		\bibliography{references}
	}

\clearpage
\appendix
\section*{Supplementary Material}
\addcontentsline{toc}{section}{Supplementary Material}

\setcounter{equation}{0}
\renewcommand{\theequation}{S\arabic{equation}}
\setcounter{figure}{0}
\renewcommand{\thefigure}{S\arabic{figure}}
\setcounter{table}{0}
\renewcommand{\thetable}{S\arabic{table}}
\setcounter{algorithm}{0}
\renewcommand{\thealgorithm}{S\arabic{algorithm}}

\renewcommand{\theHequation}{supp.\arabic{equation}}
\renewcommand{\theHfigure}{supp.\arabic{figure}}
\renewcommand{\theHtable}{supp.\arabic{table}}
\providecommand{\theHalgorithm}{}
\renewcommand{\theHalgorithm}{supp.\arabic{algorithm}}

\section{Scaling to Vision Transformer Backbones}
To demonstrate the generalizability of QueenVIS beyond standard convolutional networks, we evaluate our query enrichment framework using the Swin-L (Swin Transformer Large) backbone~\cite{liu2021swin}. Table~\ref{tab:swin_comparison} presents the comparison of QueenVIS against recent state-of-the-art methods on the YouTube-VIS 2019, YouTube-VIS 2021, and OVIS datasets. 

The integration of ViT-based features into our image-only training pipeline requires no architectural modifications. The dense appearance and spatial priors extracted by our auxiliary branches seamlessly translate to transformer-based feature maps. As shown in Table~\ref{tab:swin_comparison}, QueenVIS with a Swin-L backbone consistently improves the image-only MinVIS baseline, with a +4.5 AP gain on YouTube-VIS 2021 and +1.6 AP on both YouTube-VIS 2019 and OVIS. Despite relying solely on image-level supervision, it surpasses several video-supervised methods, including SeqFormer and Mask2Former-VIS, on all three benchmarks, and narrows the gap to the strongest video-supervised approaches: on YouTube-VIS 2021 it trails VISAGE, CTVIS, and CAVIS by only 0.8--1.4 AP. As in the ResNet-50 setting, a larger gap remains on the heavily occluded OVIS, where the limit of image-only training is most visible. These results indicate that our single-frame query enrichment strategy is not specific to convolutional backbones and continues to benefit from stronger, transformer-based feature extractors.

\begin{table*}[!t]
	\centering
	\caption{Comparison with state-of-the-art VIS methods using the Swin-L backbone. Metrics denote Average Precision (AP) at various IoU thresholds and Average Recall (AR). QueenVIS maintains robust performance against video-supervised methods without requiring any multi-frame training.}
	\label{tab:swin_comparison}
	\resizebox{\linewidth}{!}{
		\begin{tabular}{l c | ccccc | ccccc | ccccc}
			\toprule
			\multirow{2}{*}{Method} & \multirow{2}{*}{Setting} & \multicolumn{5}{c|}{OVIS} & \multicolumn{5}{c|}{YouTube-VIS 2019} & \multicolumn{5}{c}{YouTube-VIS 2021} \\
			\cmidrule(lr){3-7} \cmidrule(lr){8-12} \cmidrule(lr){13-17}
			& & AP & AP$_{50}$ & AP$_{75}$ & AR$_1$ & AR$_{10}$ & AP & AP$_{50}$ & AP$_{75}$ & AR$_1$ & AR$_{10}$ & AP & AP$_{50}$ & AP$_{75}$ & AR$_1$ & AR$_{10}$ \\
			\midrule
			\multicolumn{17}{l}{\textit{Video Supervision}}\\
			\midrule
			SeqFormer~\cite{wu2022seqformer} & Offline & 31.7 & 54.8 & 31.2 & 15.6 & 39.4 & 61.2 & 84.1 & 66.8 & 52.8 & 67.2 & 53.6 & 76.5 & 59.9 & 43.1 & 61.6 \\
			Mask2Former-VIS~\cite{cheng2021mask2former} & Offline & 35.5 & 62.4 & 35.8 & 16.9 & 42.1 & 60.4 & 84.4 & 67.0 & -- & -- & 52.6 & 76.4 & 57.2 & -- & -- \\
			VITA~\cite{heo22vita} & Offline & 44.8 & 70.9 & 46.1 & 18.7 & 48.9 & 61.5 & 85.9 & 67.7 & 54.4 & 67.6 & 57.5 & 80.6 & 61.0 & 47.7 & 62.6 \\
			GenVIS~\cite{heo2023generalized} & Offline & 45.4 & 73.1 & 45.4 & 19.3 & 49.3 & 64.0 & 87.0 & 71.3 & 55.4 & 67.7 & 60.1 & 82.9 & 66.0 & 48.2 & 64.6 \\
			VISAGE~\cite{kim2024visage} & Online & 46.2 & 71.9 & 48.5 & 19.2 & 49.7 & 64.5 & 86.8 & 72.1 & 55.3 & 68.2 & 60.6 & 83.3 & 66.8 & 49.0 & 66.0 \\
				CTVIS~\cite{ying2023ctvis} & Online & 46.9 & 71.5 & 47.5 & 19.1 & 52.1 & 65.6 & 87.7 & 72.2 & 56.5 & 70.4 & 61.2 & 84.0 & 68.8 & 48.0 & 65.8 \\
				CAVIS~\cite{lee2025cavis} & Online & 48.6 & 74.0 & 52.5 & 19.5 & 53.3 & 66.0 & 89.5 & 73.3 & 56.8 & 71.4 & 61.1 & 84.1 & 69.2 & 48.2 & 66.3 \\
			\midrule
			\multicolumn{17}{l}{\textit{Image-Only Training (No Video Supervision)}}\\
			\midrule
			MinVIS~\cite{huang2022minvis} & Online & 39.4 & 61.5 & 41.3 & 18.1 & 43.3 & 61.6 & 83.3 & \textbf{68.6} & 54.8 & 66.6 & 55.3 & 76.6 & 62.0 & 45.9 & 60.8 \\
			\textbf{QueenVIS} & Online & \textbf{41.0} & \textbf{62.2} & \textbf{43.0} & \textbf{18.4} & \textbf{45.0} & \textbf{63.2} & \textbf{85.3} & 68.3 & \textbf{55.5} & \textbf{68.4} & \textbf{59.8} & \textbf{81.5} & \textbf{65.2} & \textbf{48.3} & \textbf{64.2} \\
			\bottomrule
		\end{tabular}
	}
\end{table*}

\section{Inference Efficiency}
As detailed in the main manuscript, our auxiliary feature and center prediction heads are entirely discarded post-training, meaning QueenVIS introduces zero additional parameters during inference. The only computational overhead stems from our online tracking components: the query propagation and memory-bank matching schemes, which operate sequentially across frames.

Table~\ref{tab:inference_fps} quantifies this runtime trade-off against the MinVIS baseline. All measurements were conducted on a single NVIDIA H200 GPU using the ResNet-50 backbone. The throughput gap stems entirely from our additional online tracking components, namely query propagation and memory-bank matching, which run sequentially on top of the per-frame inference shared with MinVIS. This overhead reduces throughput from $\sim$27 to $\sim$20 FPS, which remains well-suited for real-time applications and preserves causal temporal awareness while circumventing the memory footprint and computational burden of optimizing long video clips during training.

\begin{table}[t]
	\centering
	\caption{Inference speed and AP comparison on YouTube-VIS 2021 (ResNet-50) evaluated on a single NVIDIA H200.}
	\label{tab:inference_fps}
	\setlength{\tabcolsep}{7pt}
	\renewcommand{\arraystretch}{1.1}
	\resizebox{\linewidth}{!}{%
	\begin{tabular}{lcccc}
		\toprule
		Method & AP$\uparrow$ & FPS$\uparrow$ & GFLOPs & \#Params\\
		\midrule
		MinVIS~\cite{huang2022minvis} & 44.2 & \textbf{26.92} & 266.09 & 293.89\,M \\
		\textbf{QueenVIS} & \textbf{50.9} & 20.02 & 266.09 & 300.34\,M \\
		\bottomrule
	\end{tabular}
	}
\end{table}

\begin{figure*}[t]
	\centering
	\includegraphics[width=0.95\linewidth]{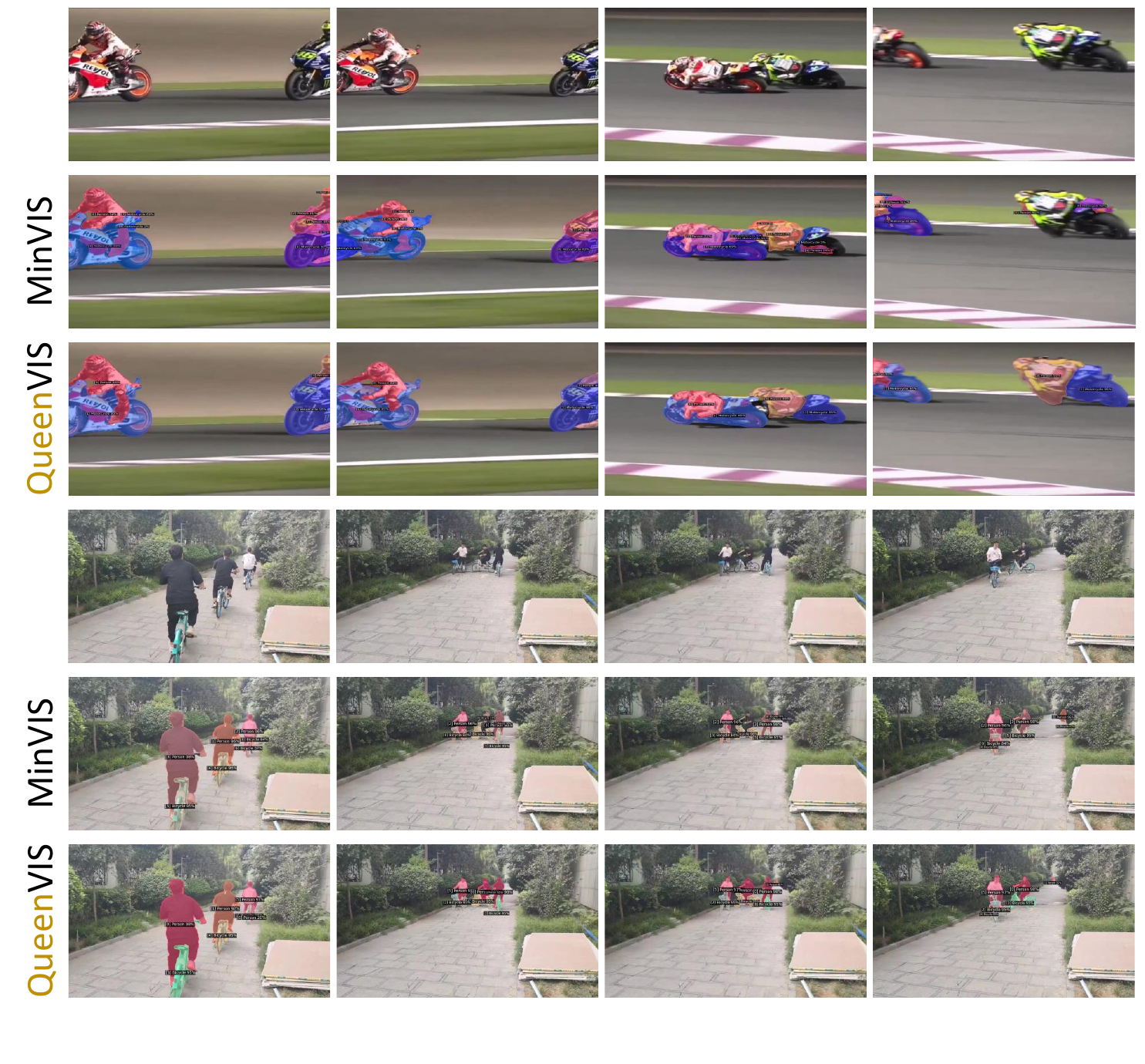}
	\caption{\textbf{Additional qualitative comparison between MinVIS and QueenVIS on OVIS.} For each sequence, rows show, from top to bottom, the input frames, the MinVIS baseline, and our QueenVIS predictions; identical instances keep a consistent color across frames. Compared with MinVIS, QueenVIS preserves object identities through occlusion and crossing trajectories and produces more complete masks, all without any video-level supervision.}
	\label{fig:supp_compare}
\end{figure*}

\section{Relational Distillation vs. Absolute Feature Regression}
\textbf{Why per-query regression, not relational distillation?} A natural alternative to our per-query feature loss is to match the \emph{relations} between queries rather than their absolute descriptors. We tested this with a feature-geometry distillation loss inspired by~\cite{park2019rkd}. Let $\mathcal{P}$ be the set of unique matched-object pairs $(i, j)$ within a frame. We compute the pairwise cosine similarity of the object query embeddings ($S^Q_{i,j}$) and of the ground-truth backbone descriptors ($S^G_{i,j}$):
\begin{equation}
	S^Q_{i,j} = \frac{\mathbf{q}_i \cdot \mathbf{q}_j}{\|\mathbf{q}_i\|_2 \|\mathbf{q}_j\|_2}, \quad S^G_{i,j} = \frac{\mathbf{f}_i \cdot \mathbf{f}_j}{\|\mathbf{f}_i\|_2 \|\mathbf{f}_j\|_2}
\end{equation}
The geometric distillation objective penalizes the structural divergence between the query space and the target feature space by minimizing the $\ell_1$ distance between these similarity distributions:
\begin{equation}
	\mathcal{L}_{\text{geom}} = \frac{1}{|\mathcal{P}|} \sum_{(i,j) \in \mathcal{P}} \left| S^Q_{i,j} - S^G_{i,j} \right|
\end{equation}
Under a matched base (query propagation and memory bank, without center prediction), replacing our absolute per-query regression ($\mathcal{L}_{\text{feat}}$) with this relational term degrades performance: AP drops from $49.6$ (FP) to $47.5$ (FA), a $2.1$ AP loss (Table~\ref{tab:ablation_alignment}). We tuned $\lambda_{\text{geom}}$ and report its best value ($0.1$); neither smaller nor larger weights closed the gap, so it does not stem from loss weighting. A likely cause is a mismatch between what the relational term supervises and how queries are used at inference: the loss constrains only the \emph{relative} pairwise geometry among instances and leaves each query's absolute embedding under-determined, whereas association matches queries by their absolute embeddings through cosine similarity in the matcher and memory bank. Absolute regression supplies precisely this per-query target. The effect is compounded by the few objects typically present per frame, which makes the set of matched pairs small and the relational signal comparatively sparse and noisy.

\begin{table}[tb]
	\centering
	\caption{Absolute feature regression (FP) vs.\ relational distillation (FA) on YouTube-VIS 2021. Under a matched base (QP\,+\,MB, no CP), replacing our per-query regression with the relational term lowers AP by $2.1$, supporting absolute feature regression. FP: Feature Prediction loss ($\mathcal{L}_{\text{feat}}$); FA: relational geometric distillation ($\mathcal{L}_{\text{geom}}$); CP: Center Prediction; QP: Query Propagation; MB: Memory Bank.}
	\label{tab:ablation_alignment}
	\setlength{\tabcolsep}{4pt}
	\resizebox{\linewidth}{!}{%
	\begin{tabular}{lccc}
		\toprule
		Configuration & AP & AP$_{50}$ & AP$_{75}$ \\
		\midrule
		Final Model (FP + CP + QP + MB) & \textbf{50.9} & \textbf{72.1} & \textbf{55.8} \\
		FP ($\lambda_{\text{feat}}=0.2$) + QP + MB & 49.6 & 71.1 & 54.7 \\
		FA ($\lambda_{\text{geom}}=0.1$) + QP + MB & 47.5 & 68.8 & 52.5 \\
		\bottomrule
	\end{tabular}
	}
\end{table}

\section{Sensitivity to Auxiliary Loss Weights}
Our training objective augments the MinVIS baseline with two auxiliary terms, weighted by $\lambda_{\text{feat}}$ (feature prediction, FP) and $\lambda_{\text{center}}$ (center prediction, CP). Table~\ref{tab:queenvis_supp} reports a sensitivity sweep over both on YouTube-VIS 2021; all configurations share the same training-free association (query propagation and memory bank) at inference, so differences reflect the loss weighting alone.

We first vary the feature-prediction weight in isolation. Overly large weights let the auxiliary regression dominate the classification and mask losses and degrade accuracy: $\lambda_{\text{feat}}=4$ and $\lambda_{\text{feat}}=2$ reach only $47.4$ and $48.4$ AP, whereas $\lambda_{\text{feat}}=0.2$ is clearly best at $49.6$ AP. Fixing $\lambda_{\text{feat}}=0.2$, we then add center prediction. The model is robust to $\lambda_{\text{center}}$: every tested value improves over the FP-only model, and $\lambda_{\text{center}}=0.05$ yields the best overall result ($50.9$ AP), our final QueenVIS configuration. The narrow spread across center weights indicates that the gain comes from the presence of the spatial objective rather than from its precise weighting.

\begin{table*}[tb]
	\centering
	\caption{Sensitivity to the auxiliary loss weights $\lambda_{\text{feat}}$ (feature prediction, FP) and $\lambda_{\text{center}}$ (center prediction, CP) on YouTube-VIS 2021 (ResNet-50). All configurations use query propagation and a memory bank at inference. Best results in \textbf{bold}.}
	\label{tab:queenvis_supp}
	\resizebox{0.75\linewidth}{!}{
		\begin{tabular}{lccccc}
			\toprule
			Configuration & AP & AP$_{50}$ & AP$_{75}$ & AR$_1$ & AR$_{10}$ \\
			\midrule
			FP ($\lambda_{\text{feat}}=4$)   & 47.4 & 69.1 & 52.5 & 42.2 & 55.3 \\
			FP ($\lambda_{\text{feat}}=2$)   & 48.4 & 70.5 & 52.3 & 41.7 & 55.4 \\
			FP ($\lambda_{\text{feat}}=0.2$) & 49.6 & 71.1 & 54.7 & 42.8 & 56.2 \\
			FP ($\lambda_{\text{feat}}=0.2$) + CP ($\lambda_{\text{center}}=0.1$)  & 49.7 & 70.9 & 55.0 & \textbf{43.0} & 56.9 \\
			FP ($\lambda_{\text{feat}}=0.2$) + CP ($\lambda_{\text{center}}=0.01$) & 49.8 & 70.4 & 54.7 & 42.7 & 56.7 \\
			\textbf{FP ($\lambda_{\text{feat}}=0.2$) + CP ($\lambda_{\text{center}}=0.05$) (QueenVIS)} & \textbf{50.9} & \textbf{72.1} & \textbf{55.8} & 42.8 & \textbf{57.3} \\
			\bottomrule
		\end{tabular}
	}
\end{table*}

\section{Additional Qualitative Results}
\begin{figure*}[!t]
	\centering
	\includegraphics[width=0.95\linewidth]{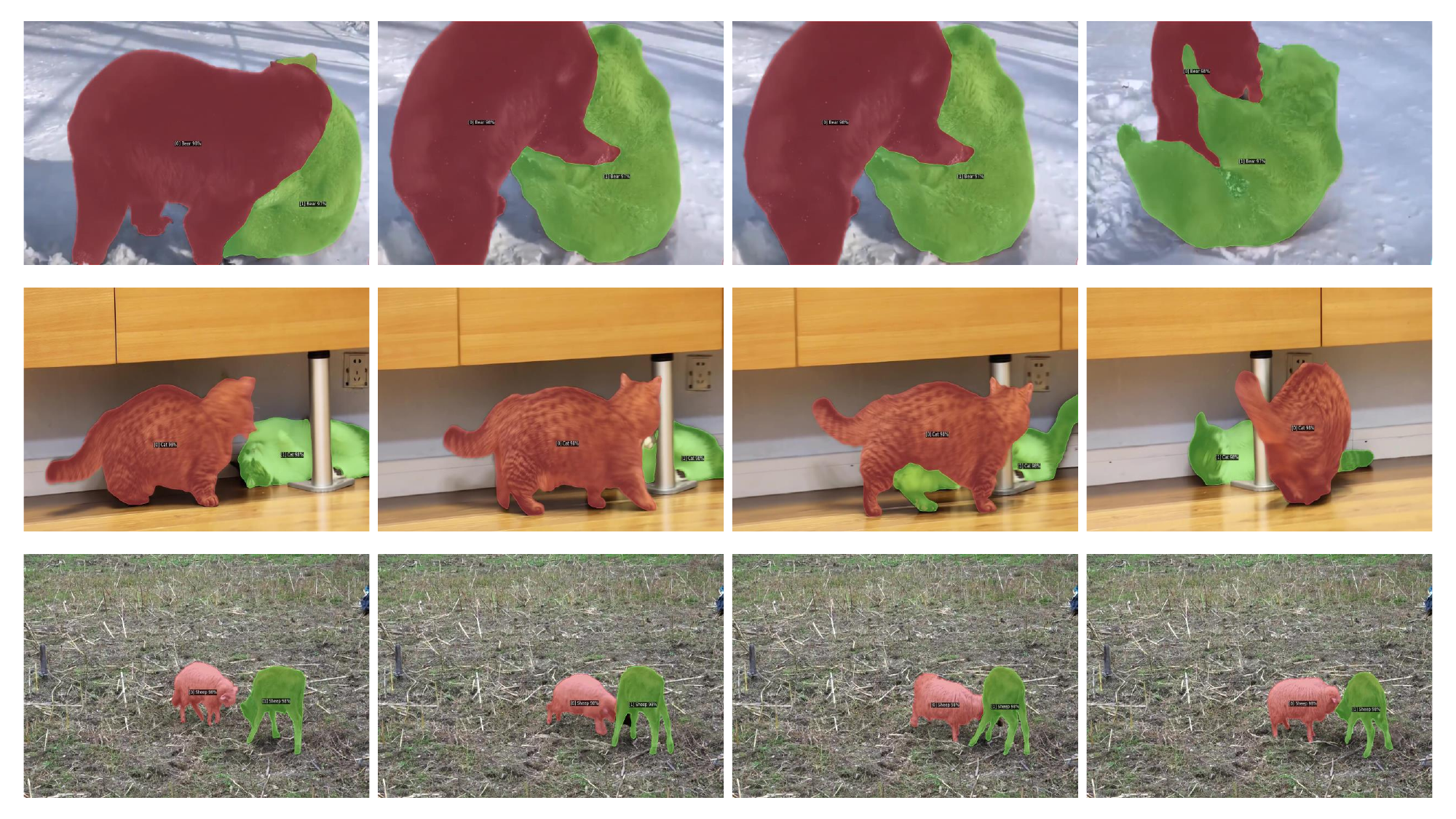}
	\caption{\textbf{Additional QueenVIS results on OVIS.} Qualitative tracking results of QueenVIS on three OVIS sequences. Consistent colors denote consistent identities maintained across the temporal window under rapid motion, scale variation, and occlusion, illustrating that query-space appearance and spatial enrichment stabilizes inference without relying on video-level ground truth.}
	\label{fig:supp_ours}
\end{figure*}

To further demonstrate the robustness of QueenVIS, we provide additional qualitative visualizations. Figure~\ref{fig:supp_compare} compares QueenVIS against the MinVIS baseline on challenging sequences: where MinVIS drifts into identity switches and fragmented masks, QueenVIS keeps identities consistent and segments objects more completely, without any video-level supervision. Figure~\ref{fig:supp_ours} further shows QueenVIS predictions on three additional sequences. Together, these results suggest that our query enrichment strategies successfully retain tracking consistency across challenging sequences involving rapid motion, severe scale variation, and intersection of visually similar instances.

\makeatletter
\setlength{\@dblfptop}{0pt}
\makeatother

\end{document}